 \documentclass[pmlr,twocolumn,10pt,breaklinks, x11names,table]{jmlr} 

 \let\SUP\textsuperscript

\usepackage{url}
\usepackage{breakurl}
\usepackage{ragged2e}

\usepackage{booktabs}
\usepackage[load-configurations=version-1]{siunitx} 

\theorembodyfont{\upshape}
\theoremheaderfont{\scshape}
\theorempostheader{:}
\theoremsep{\newline}

\jmlrvolume{ML4H Extended Abstract Arxiv Index}
\jmlryear{2021}
\jmlrsubmitted{2021}
\jmlrpublished{}
\jmlrworkshop{Machine Learning for Health (ML4H) 2021} 

\title[Machine Learning for Diagnosis of Parkinson's Disease]{Machine Learning for Real-Time, Automatic, and Early Diagnosis of Parkinson’s Disease by Extracting Signs of Micrographia from Handwriting Images}

\author{Riya Tyagi \nametag{\thanks{Authors contributed equally}},
Tanish Tyagi \nametag{\footnotemark[1]}, 
Ming Wang,
Lijun Zhang
\centering \Email{ 
\\[\bigskipamount]  
rtyagi@exeter.edu, ttyagi@mgh.harvard.edu, mwang@phs.psu.edu, lzhang6@pennstathealth.psu.edu}
}

\begin{document}

\maketitle

\begin{abstract}
Parkinson’s disease (PD) is debilitating, progressive, and clinically marked by motor symptoms. As the second most common neurodegenerative disease in the world, it affects over 10 million lives globally. Existing diagnoses methods have limitations, such as the expense of visiting doctors and the challenge of automated early detection, considering that behavioral differences in patients and healthy individuals are often indistinguishable in the early stages. However, micrographia, a handwriting disorder that leads to abnormally small handwriting, tremors, dystonia, and slow movement in the hands and fingers, is commonly observed in the early stages of PD. In this work, we apply machine learning techniques to extract signs of micrographia from drawing samples gathered from two open-source datasets and achieve a predictive accuracy of 94\%. This work also sets the foundations for a publicly available and user-friendly web portal that anyone with access to a pen, printer, and phone can use for early PD detection.
\end{abstract}

\begin{keywords}
Parkinson’s Disease, Machine Learning, Micrographia, Early Detection, Predictive accuracy
\end{keywords}

\section{Introduction}
\label{sec:intro}

Parkinson’s disease (PD) is a neurodegenerative, chronic, and progressive nervous system disorder that affects movement. PD affects more than 10 million people worldwide, and this number is expected to double by 2030. \citep{choi2017refining} Symptoms include tremors, slowness of movement, stiffness, and changes in writing skills. PD occurs when dopamine-producing nerve cells die, a process that can take almost 10 years to reach its late and most severe stages. A lack of dopamine results in further critical symptoms, such as depression, anxiety, sleep disturbances, and dementia. 

Parkinson's disease cannot be treated, although specific drugs and medications can assist with early-stage symptoms, potentially preventing the progression to more severe stages. Early detection and prognosis of PD are crucial for assisting patients to retain a good quality of life. However, diagnosing Parkinson’s disease in its early stages is a very challenging task, and there are currently no specific tests designed or approved to diagnose PD \citep{national_institute_of_neurological_disorders_and_stroke}. Diagnosis is based on a thorough examination performed by a trained medical official, including a neurological examination, medical history evaluation, blood and laboratory tests, and brain scans. Even with these procedures, PD is misdiagnosed up to 30\% of the time because of the many PD mimics, such as Essential Tremors and Drug-Induced Parkinson’s. Additionally, up to 20\% of PD patients are undiagnosed \citep{schrag_ben-shlomo_quinn_2002}. Receiving a PD diagnosis is expensive, time-consuming, and non-accessible, as reported by 21\% of patients who had to visit their general provider thrice before receiving a specialist referral for their condition. \citep{gavidia_2020}

Tools that make the diagnosis process accurate, accessible, automatic, real-time, early, and free of cost can have a tremendous impact on Parkinson’s patients. In this study, we apply machine learning strategies to predict PD based upon drawing samples that exhibit signs of micrographia. 

\section{Related Works}
\label{related-works}
Prior works have incorporated voice recordings, electroencephalogram (EEG) signals, and smart devices to create models that diagnose PD. \citet{zhao2014automatic} hypothesized that patients with PD exhibit deficits in the production of emotional speech. To test this conjecture, voice recordings from five patients and seven healthy individuals were used to detect Parkinson’s disease. Naive Bayes, Random Forests, and Support Vector Machines were applied for classification, achieving 65.5\% and 73.33\% accuracies on classifying PD and control. \citet{oh2020deep} utilized EEG signals and a Convolutional Neural Network (CNN) to detect PD by assessing whether the EEG signals depicted brain abnormalities, achieving 88.25\% accuracy. \citet{drotar2014analysis} employed the use of a digitizing tablet to assess both in-air and on-surface kinematic variables during handwriting of a sentence in 37 PD patients on medication and 38 age and gender-matched healthy controls. Using Support Vector Machines, an accuracy of 85.61\% was achieved. 

These studies require patients to visit a neurologist to access a smart device and undergo various scans like EEG and PET, making the process expensive, time-consuming, and difficult to access. In this research, we aim to make the PD diagnosis more accurate, efficient, and accessible by using images of handwriting samples to automate the diagnosis process. 

\section{Dataset and Preprocessing}
\label{sec:Dataset+Preprocessing}

\paragraph{Dataset}
\label{sec:Dataset} We formed our dataset by combining the New HandPD and Old HandPD dataset \citep{pereira2016new}. It consisted of 158 individuals, split into 53 healthy individuals and 105 PD patients. Each patient drew 8 images: 4 spirals and 4 meanders. With these images, we constructed 2 datasets, one with extracted numerical features and the other with images. The process of assembling the image dataset and its uses are discussed further in Appendix A.

\begin{table*}[hbtp]
\centering 
\floatconts
{tab:table1}
{\caption{Demographics of Data}} 
    {
        \begin{tabular}{lcccccc}
        \toprule
        \bfseries & \bfseries Mean Age (SD) & \bfseries Left Handed & \bfseries Right Handed & \bfseries \% Female & \bfseries Counts \\
        \midrule
        Healthy & 44.11 (15.46) & 7 & 46 & 54.7\% & 53 \\
        PD & 58.75 (7.51) & 5 & 69 & 20.3\% & 74 \\
        \bottomrule
        \end{tabular}
    }
\end{table*}

\paragraph{Preprocessing}
\label{sec:Preprocessing} 
We mimicked the feature extraction process used by \citep{pereira2016new} to extract 9 numerical features from the images. First, we extracted the handwritten trace (HT) and exam trace (ET) were extracted from each image. The HT is the outline the patient drew and the ET is the outline the patient tried to replicate. Appendix \ref{app:webportal} show images of the extracted HT and ET. Using the traces, the below features were computed:

\begin{enumerate}
\item[F1:] The root mean square (RMS) of the differences between the HT and ET radii. The radius of the HT or ET can be defined as the length of the straight line that connects an arbitrary point to the center of the HT or ET. 
\[RMS = \sqrt{\frac{1}{N} * \sum_\SUP{i=1}^N(r_\SUP{HT}^i - r_\SUP{ET}^i)^2}\]

$N$ is the number of sample points on the HT and ET, and $r_\SUP{HT}^i$ and $r_\SUP{ET}^i$ denote the HT and ET radii for the $i$th point.

\item[F2:] The maximum difference between the HT and ET radii. 
\[\Delta max = max(|r_\SUP{HT}^i - r_\SUP{ET}^i|)\]

\item[F3:] The minimum difference between the HT and ET radii. 
\[\Delta min = min(|r_\SUP{HT}^i - r_\SUP{ET}^i|)\]

\item[F4:] The standard deviation of the differences between the HT and ET radii. 
\[s = \sqrt{\frac{1}{N-1}\sum_{i=1}^N(r_\SUP{HT}^i - r_\SUP{ET}^i)^2}\]

\item[F5:] The Mean Relative Tremor (MRT) of a given individual’s HT. This is defined as the mean difference between the radius of a given sample and its D left-nearest neighbors. To find D, D = {1, 3, 5, 7, 10, 15, 20} were tested, with D = 10 maximizing the PD detection rate. 
\[MRT = \frac{1}{N-D}\sum_\SUP{i=D}^N(|r_\SUP{HT}^\SUP{i-D+1} - r_\SUP{ET}^i|)\]

Figures 6-8 are computed utilizing the equation for relative tremor: $|r_\SUP{ET}^i - r_\SUP{HT}^\SUP{i-D+1}|$
\item[F6:] Maximum ET
\item[F7:] Minimum ET
\item[F8:] Standard Deviation of ET values
\item[F9:] The number of times the difference between HT and ET radii changes from positive to negative.
\end{enumerate}

The features were then normalized as follows:
\[f[i]' = \sum_\SUP{i=1}^9\frac{f[i] - avg(i)}{std(i)}\]
f[i]' represents the normalized equivalent of feature f[i] for features 1-9.

Two other features we include in our predictions are a patient's gender and age, considering that Parkinson's disease mainly impacts individuals above age 50 and is 1.5 times more common in males \citep{reekes2020sex}. 

Additionally, we noticed there was a large class imbalance in our dataset, with 53 healthy and 105 PD patients. After performing experiments with our machine learning models, we discovered that this imbalance led to overfitting and undesirable performances, even after weighting the classes. To cope with this, we truncated our original dataset by removing the 31 PD patients from the New HandPD dataset, which greatly boosted our model's ability to learn useful patterns. Our final dataset of 127 patients, 53 healthy and 74 PD, was randomly split between train (76.5\%), validation (8.5\%), and hold-out test (15\%) sets. Table \ref{tab:table1} shows the demographics for our cohort of individuals. 

\begin{table*}[hbtp]
\centering 
\floatconts
{tab:table2}
{\caption{Model Performance}}
    {
        \begin{tabular}{lcccccccccc}
        \toprule
        \bfseries Model & \bfseries ACC & \bfseries AUC & \bfseries FP & \bfseries FN & \bfseries Sensitivity & \bfseries Specificity & \bfseries PPV & \bfseries NPV & \bfseries Threshold \\
        \midrule
        SVM & 0.92 & 0.93 & 12 & 0 & 1.0 & 0.81 & 0.89 & 1.0 & 0.65 \\
        Logistic Regression & 0.93 & 0.96 & 3 & 7 & 0.92 & 0.95 & 0.90 & 0.96 & 0.62 \\
        \bottomrule
        \end{tabular}
    }
\end{table*}

\begin{figure*}[hbtp]
\centering
\floatconts
{fig:fig1}
{\caption{Regularized Logistic Regression (left) and Support Vector Machines (right) ROC Curves}}

    {{\includegraphics[scale = 0.55]{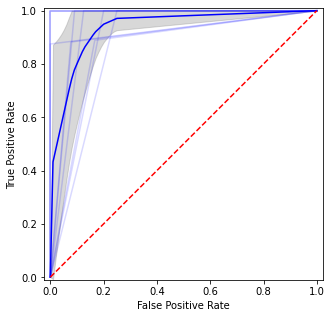} }}%
    {\includegraphics[scale = 0.55]{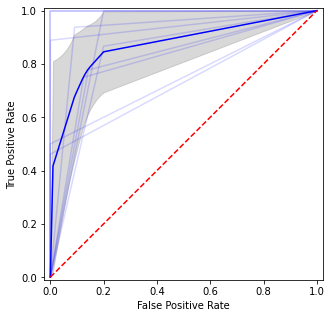}} %

\end{figure*}
\section{Methods}
\label{Methodology}

\label{sec:SVM}  
\paragraph{(1) Support Vector Machines} In model 1, we utilized a Support Vector Machine (SVM) \citep{708428} to diagnose PD given the nine extracted numerical features, age, and gender. We used 10-fold Cross-Validation (CV) \citep{stone1974cross} on the validation set to maximize accuracy by choosing an optimal kernel (Radial Base, Linear, or Polynomial), class weight (None or Balanced), probability threshold, and lambda. After training and validating the model on the image level, we tested it on the patient level, extracting all 8 images from 11 PD and 8 healthy patients (15\% test set) and selecting an optimal aggregation scheme for patient-level predictions.  

\label{sec:LogReg}  
\paragraph{(2) Regularized Logistic Regression} In model 2, we employed Regularized Logistic Regression \citep{tibshirani1996regression} strategies, regressing the PD diagnosis label against the nine extracted numerical features, age, and gender. With 10-fold CV, we iterated over different lambda values, probability thresholds, and loss and solver functions to maximize accuracy for our image-level predictions. We tested the model on the patient level, extracting all 8 images from 11 PD and 8 healthy patients (15\% test set) and determining an optimal aggregation scheme for patient-level predictions.  

\section{Results}
\label{Results} To maximize the performance of both models, we utilized a probability threshold. First, we trained, validated, and tested the models on the image level. All performance metrics in Table \ref{tab:table2} are based on image level class assignments from the held-out test set.

However, our web portal will consider 6 - 8 different drawings per patient in the diagnosis process. To mimic this functionality, we tested each model on the patient level with an aggregation scheme. We chose an optimal scheme from three options: applying a PD label to a patient if the model diagnosed 2/8 images as signs of PD, if the averaged probability score from all 8 images was greater than 0.5, or if the model predicted more than half the images to belong to a PD patient. The last scheme performed best because the model mainly predicted only a few outlier images incorrectly (if any images were predicted wrong) per patient.

Both models received the same patient-level accuracy: 0.9444. Additionally, we compared their Receiver Operating Characteristic (ROC) curves (Figure \ref{fig:fig1}) when tested on 10 random selections of 19 patients (11 PD and 8 Control). Both models performed similarly on each test set, although Regularized Logistic Regression was more consistent. Optimal hyperparameters for each model are listed below.

Support Vector Machines:
\begin{itemize}
\item kernel = ‘rbf’, transforms radial data so it can be separated with a linear separation boundary
\item class\_weight = ‘balanced’, adds weight to each class depending on the class balance
\item C = 100, lambda value
\end{itemize}

Regularized Logistic Regression:
\begin{itemize}
\item penalty = ‘elasticnet’, a combination of L1 and L2 Regularization
\item l1\_ratio = 0.75, uses 75\% L1 Regularization, 25\% L2 Regularization
\item solver = ‘saga’, only solver which supports ‘elasticnet’ penalty, the best-performer
\item C = 0.1, lambda value
\item class\_weight = ‘balanced’, adds weight to each class depending on the class balance
\item fit\_intercept = False, bias added to the decision function
\end{itemize}

\section{Conclusion and Future Works}
\label{Conclusion+FutureWorks} 
In this work, we applied classification algorithms to handwriting samples to diagnose Parkinson's disease in its early stages. Our dataset was relatively small in size, which we will address in the future by gathering data through our web portal. Our mockups and plans for the website are detailed in Appendix \ref{app:webportal}.  Additionally, we devised a handwriting assessment based on precedents and open-source datasets in this research area, which is available in Appendix \ref{app:assessment}. We strive to boost the accessibility of our handwriting assessment by online deployment through a web portal. Ideally, a patient with a touch-screen device will be able to take a quick, easy, online assessment. Once the web portal is complete, we hope to perform clinical studies with our partners from Penn State University. In these studies, we will evaluate the handwriting assessment and develop a handwriting dataset that considers temporal stages of Parkinson’s disease by analyzing handwriting and drawing samples from different stages of PD. 

The next iteration of our work with machine learning and deep learning models will be to design a Convolutional Neural Network (CNN) for analysis of the drawing images. We have already designed the feature extraction process, pictured in Appendix \ref{app:imgextraction}. The CNN will further improve prediction accuracy and work toward automatic and real-time detection of Parkinson’s disease. 

\clearpage


\bibliography{jmlr-sample}
\nocite{*}
\justify{}

\clearpage
\appendix

\section{Pictures and Use Cases of PD Diagnosis Web Portal Interface}
\label{app:webportal}

There are two use cases of the web portal: submitting handwriting assessments and viewing assessment history. The process to perform these actions is detailed below.

Submitting Handwriting Assessments

\begin{enumerate}
\item The user logs into their account on the portal (or creates an account if she does not have one) and navigates to the “Submit A Test” page.
\item The user prints out the handwriting assessment .pdf available on the page.
\item 
The user fills out the exam, scans it, and uploads a photo to the page.
\item After a few moments, the user receives a PD diagnosis, with their \% chance of having PD. The exam and its results are saved in the “Exam History” page.
\end{enumerate}

\begin{figure}[h]
\label{fig:fig6}
\centering 
\includegraphics[scale = 0.3]{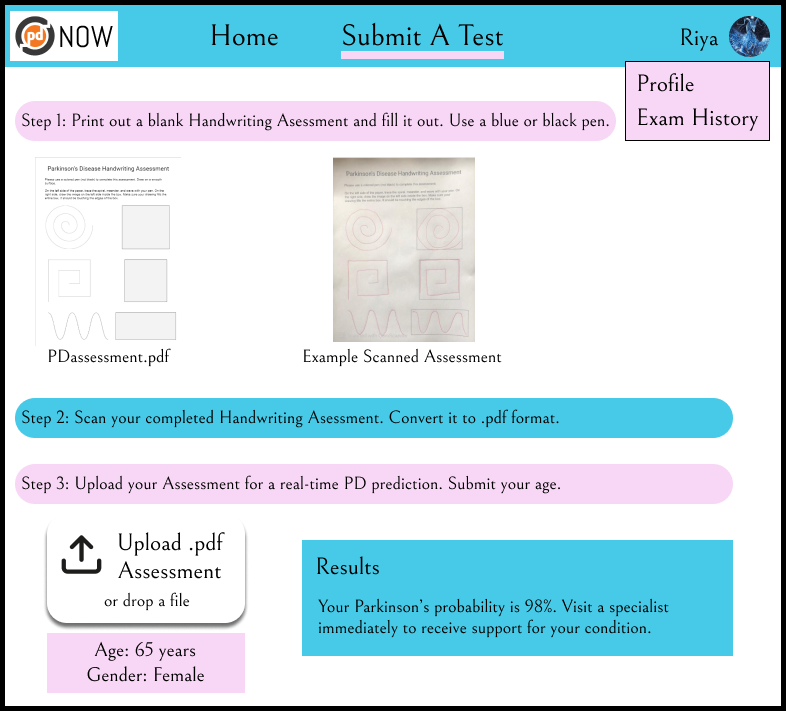}
\caption{"Submit A Test" page UI}
\end{figure}

\vfill\eject

Viewing Assessment History

\begin{enumerate}
\item The user logs into their account on the portal and navigates to the “Exam History” page.
\item The user can now view information about all their past exams, including the date, a picture of their assessment, age and gender information submitted alongside the assessment, and results.
\item The user views their previous assessment submissions by clicking on the exam icons available next to each exam number on the page.
\end{enumerate}

\begin{figure} [h]
\label{fig:fig7}
\centering 
\includegraphics[scale = 0.3]{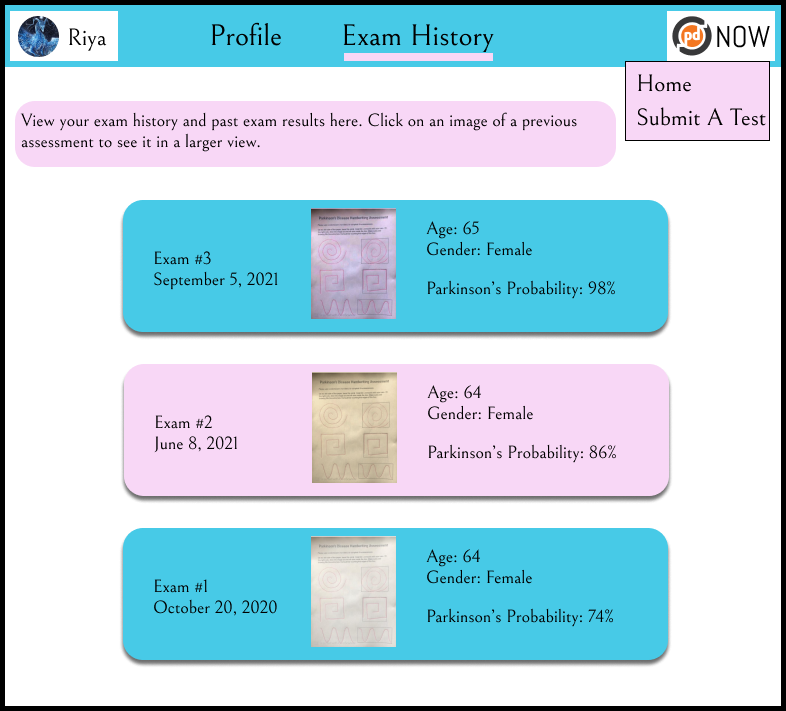}
\caption{"Exam History" page UI}
\end{figure}

\clearpage

\section{Handwriting Assessment}
\label{app:assessment}

\begin{figure} [h]
\label{fig:fig5}
\centering 
\includegraphics[scale = 0.4]{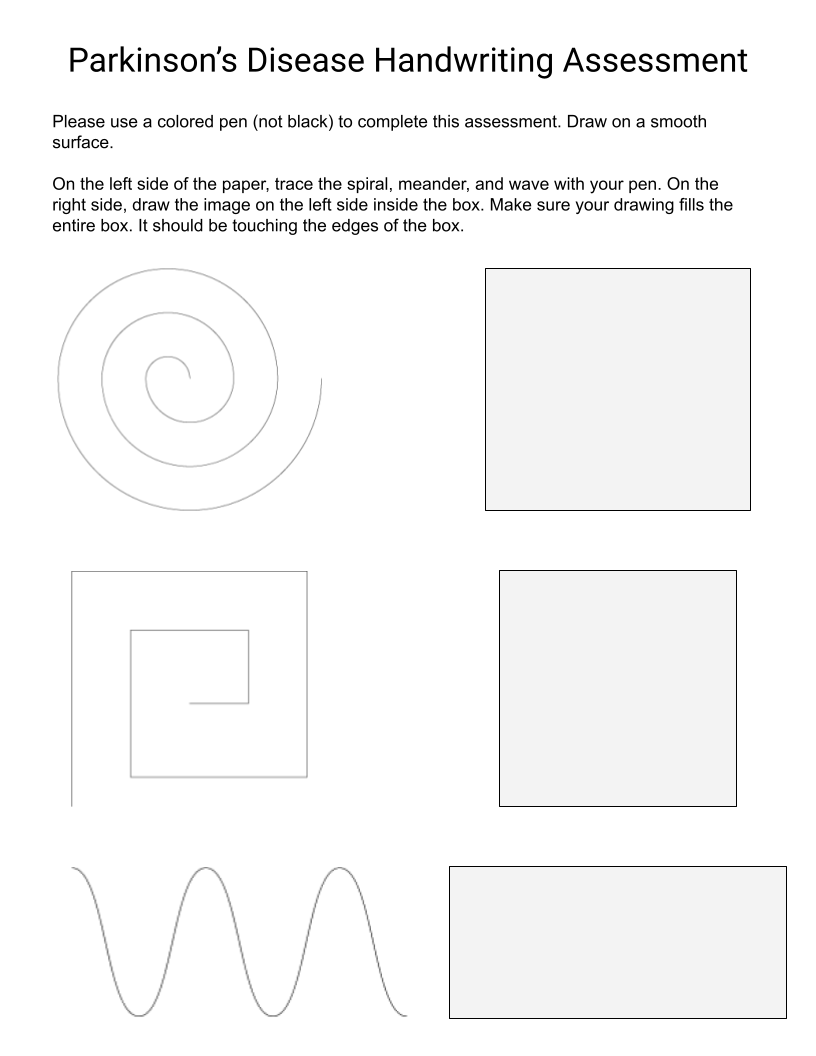}
\caption{Handwriting Assessment Developed for the Web Portal}
\end{figure}

\clearpage

\section{Image Feature Extraction Procedure}
\label{app:imgextraction}

To develop useful image features to input into a Convolutional Neural Network (CNN) model, we extracted an exam trace (ET) and a handwriting trace (HT) from all our images and combined them. To do so, we followed the process below.

Exam Trace:

\begin{enumerate}
\item Mean Blur (5 x 5 kernel)
\item Binary Threshold

I represents each pixel (x, y) in an image, with three color channels: red, green, and blue, written as I(r), I(g), and I(b)

\[I((r, g, b)) = \begin{cases} 
    0 & \text{if } I(r) < 90 \land I(g) < 90 \land I(b) < 90 \\
    1 & \text{otherwise}
\end{cases}\]

\item Dilate (4 x 4 kernel)
\end{enumerate}

\begin{figure}[h] 
\label{fig: fig2}
\centering 
\includegraphics[scale = 0.18]{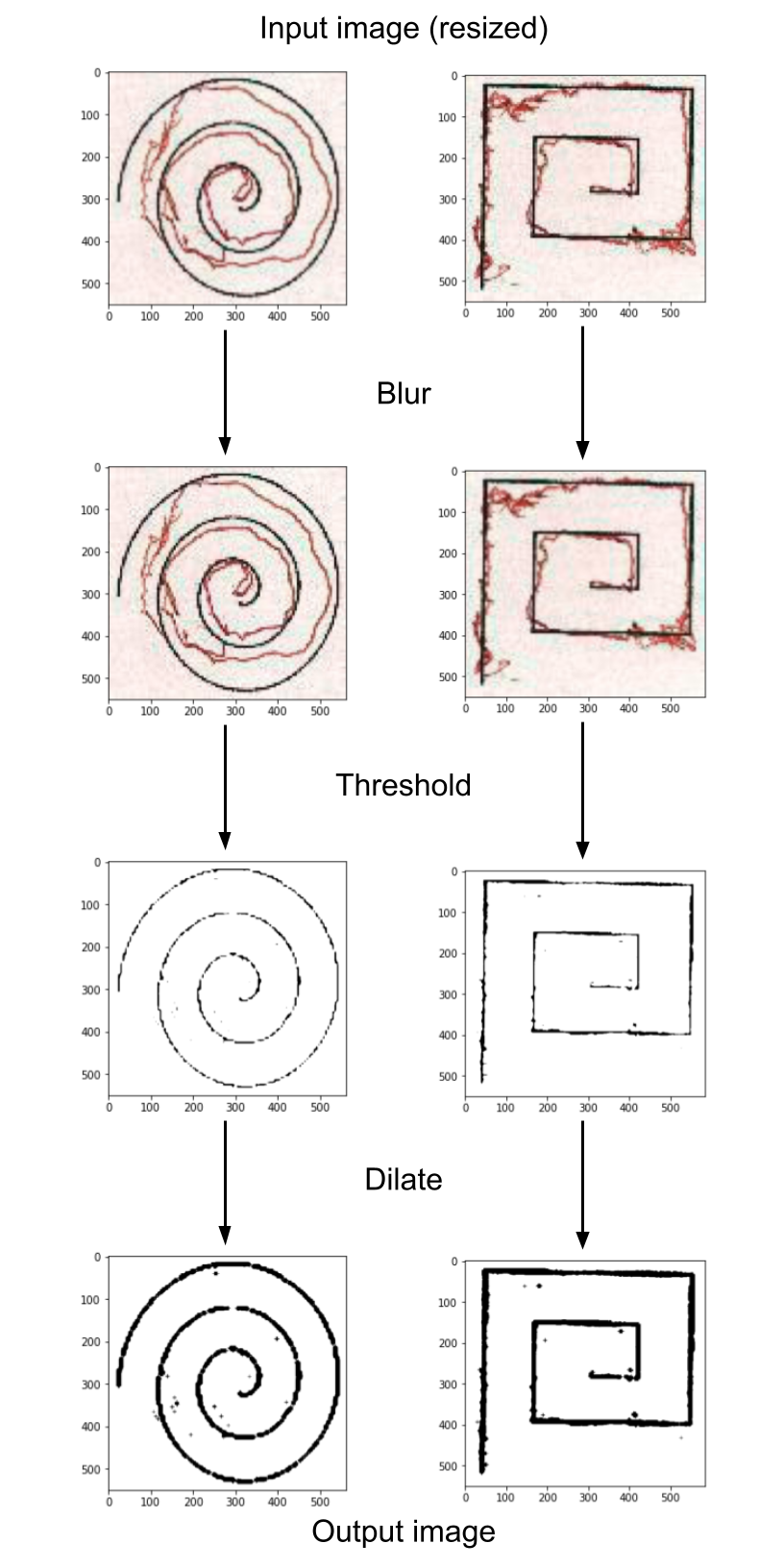}
\caption{Exam Trace Extraction Process}
\end{figure}

\vfill\eject

Handwriting Trace:

\begin{enumerate}
\item Median Blur (5 x 5 kernel)
\item Binary Threshold

Consider I to represent a pixel in an image with three color channels: red, green, and blue, written as I(r), I(g), and I(b)

\[I((r, g, b)) = \begin{cases} 
    0 & \text{if } I(r) < 200 \land I(g) < 200 \land I(b) < 200 \\
    1 & \text{otherwise}
\end{cases}\]

\item Difference Image (with extracted ET)
\item Invert and Dilate Image
\end{enumerate}

\begin{figure}[h] 
\label{fig: fig3}
\centering 
\includegraphics[scale = 0.22]{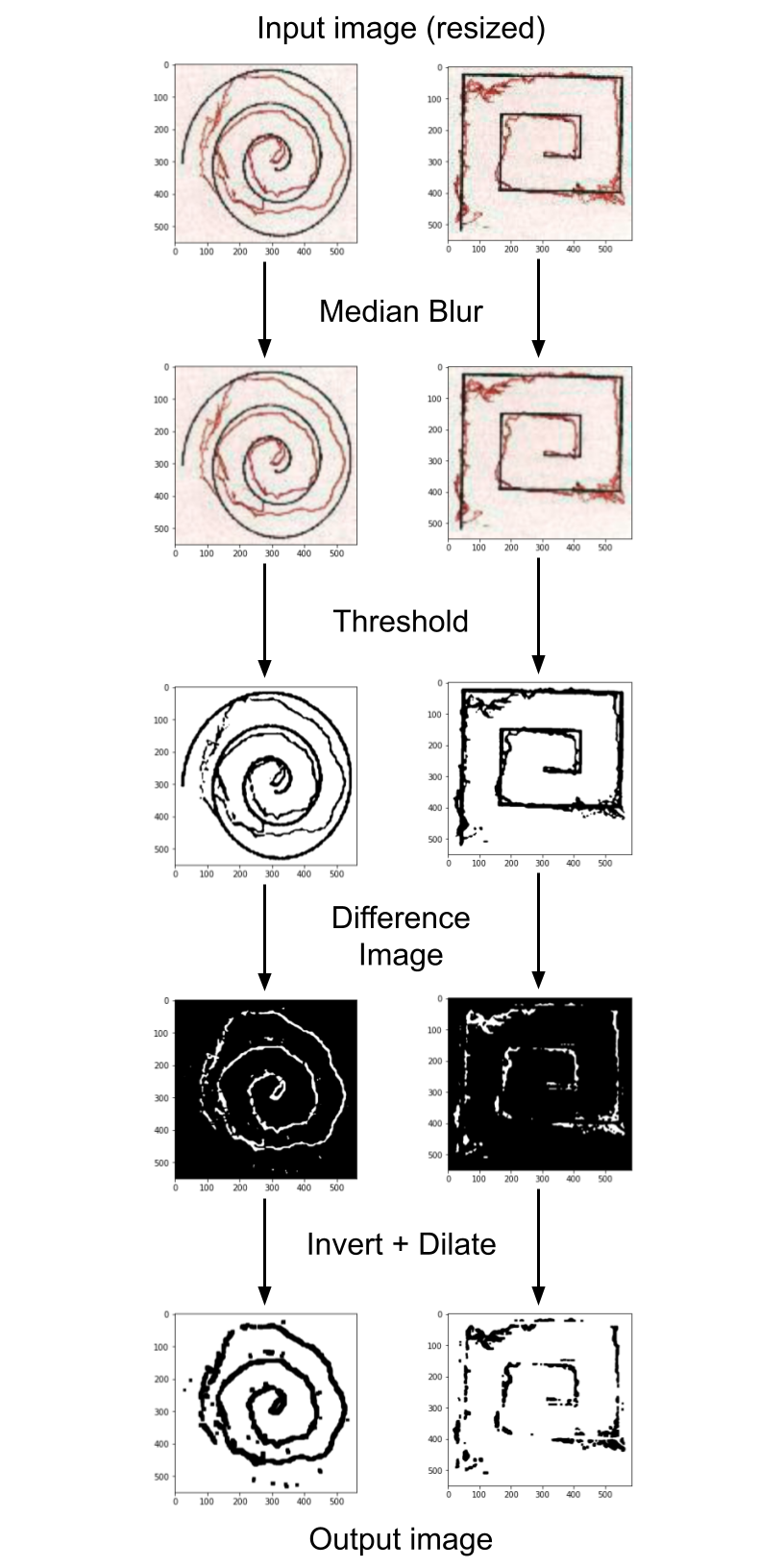}
\caption{Handwriting Trace Extraction Process}
\end{figure}

\clearpage

Next, we combined the extracted exam trace and handwriting trace, producing an image with minimal features that has the potential to form a strong CNN model.

\begin{figure} [h]
\label{fig:fig4}
\centering 
\includegraphics[scale = 0.26]{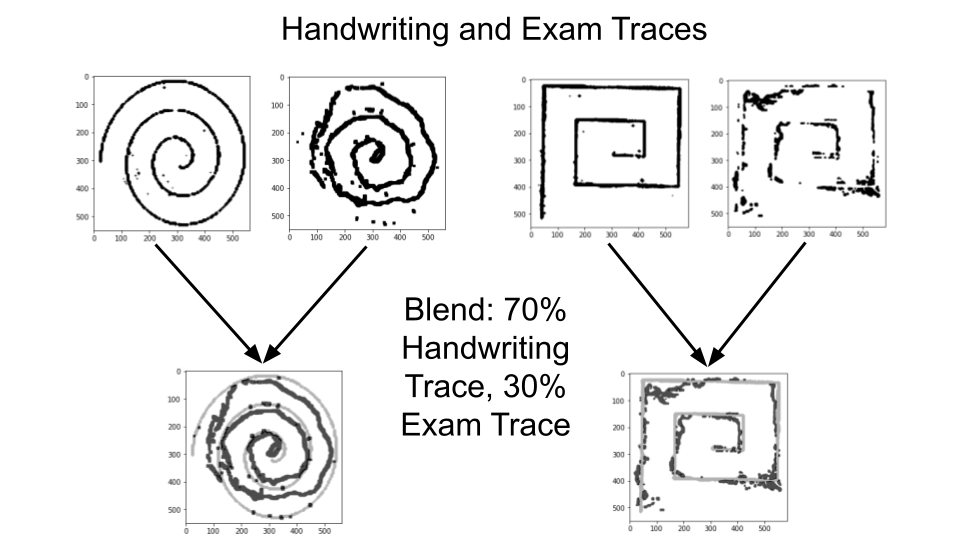}
\caption{Blending HT and ET to Form the Final Image}
\end{figure}

\end{document}